\def\year{2020}\relax
\pdfoutput=1
\documentclass[letterpaper]{article} 
\usepackage{aaai20}  
\usepackage{times}  
\usepackage{helvet} 
\usepackage{courier}  
\usepackage[hyphens]{url}  
\def\UrlFont{\rm}  
\usepackage{graphicx}
\usepackage{hhline}
\usepackage{booktabs}
\usepackage{multirow}
\usepackage{subfiles}
\usepackage{footnote}

\makeatletter
    \let\@internalcite\cite
    \def\cite{\def\citeauthoryear##1##2{##1, ##2}\@internalcite}
\makeatother

\newcommand{\newcite}[1]{\citeauthor{#1} (\citeyear{#1})}
\title{Predictive Engagement: An Efficient Metric for Automatic Evaluation of Open-Domain Dialogue Systems}
\author{
Sarik Ghazarian,\textsuperscript{\rm 1}
Ralph Weischedel,\textsuperscript{\rm 1}
Aram Galstyan,\textsuperscript{\rm 1}
Nanyun Peng\textsuperscript{\rm 1} \\
\textsuperscript{\rm 1}University of Southern California / Information Sciences Institute \\
\{sarik, weisched, galstyan, npeng\}@isi.edu
}

\begin{document}
\maketitle
\begin{abstract}
User engagement is a critical metric for evaluating the quality of open-domain dialogue systems. Prior work has focused on conversation-level engagement by using heuristically constructed  features such as the number of turns and the total time of the conversation. In this paper, we investigate the possibility and efficacy of estimating utterance-level engagement and define a novel metric, {\em predictive engagement}, for automatic evaluation of open-domain dialogue systems. Our experiments demonstrate that (1) human annotators have high agreement on assessing utterance-level engagement scores; (2) conversation-level engagement scores can be predicted from properly aggregated utterance-level engagement scores. Furthermore, we show that the utterance-level engagement scores can be learned from data. These scores can be incorporated into automatic evaluation metrics for open-domain dialogue systems to improve the correlation with human judgements. This suggests that predictive engagement can be used as a real-time feedback for training better dialogue models. 

\end{abstract}

\section{Introduction}

Given recent rapid development of open-domain dialogue systems, precise evaluation metrics seem imperative. Poor correlation between word-overlap metrics and human judgements (e.g., BLEU~\cite{DBLP:conf/acl/PapineniRWZ02}, ROUGE~\cite{rouge-a-package-for-automatic-evaluation-of-summaries})   ~\cite{DBLP:conf/emnlp/LiuLSNCP16,DBLP:conf/emnlp/NovikovaDCR17} plus the expense and time demands of  human evaluations, motivate dialogue system researchers to seek better automatic evaluation metrics.  

The evaluation of open-domain dialogue systems is especially challenging. 
Recent works have proposed automatic trainable evaluation methods that focus on a specific aspect of a dialogue system's quality. \newcite{DBLP:conf/acl/LoweNSABP17} trained an evaluation model on top of a human annotated dataset to infer an appropriateness score for each response. \newcite{DBLP:conf/aaai/TaoMZY18} combined the referenced scores (the similarity of a generated response to a ground-truth response) and the unreferenced scores (the relevancy of a generated response to a given query) to obtain better correlation with human judgements. \newcite{ghazarian2019better} further improved the accuracy of such metrics by leveraging contextualized embeddings.

\begin{figure}[t]
\includegraphics[width=\linewidth]{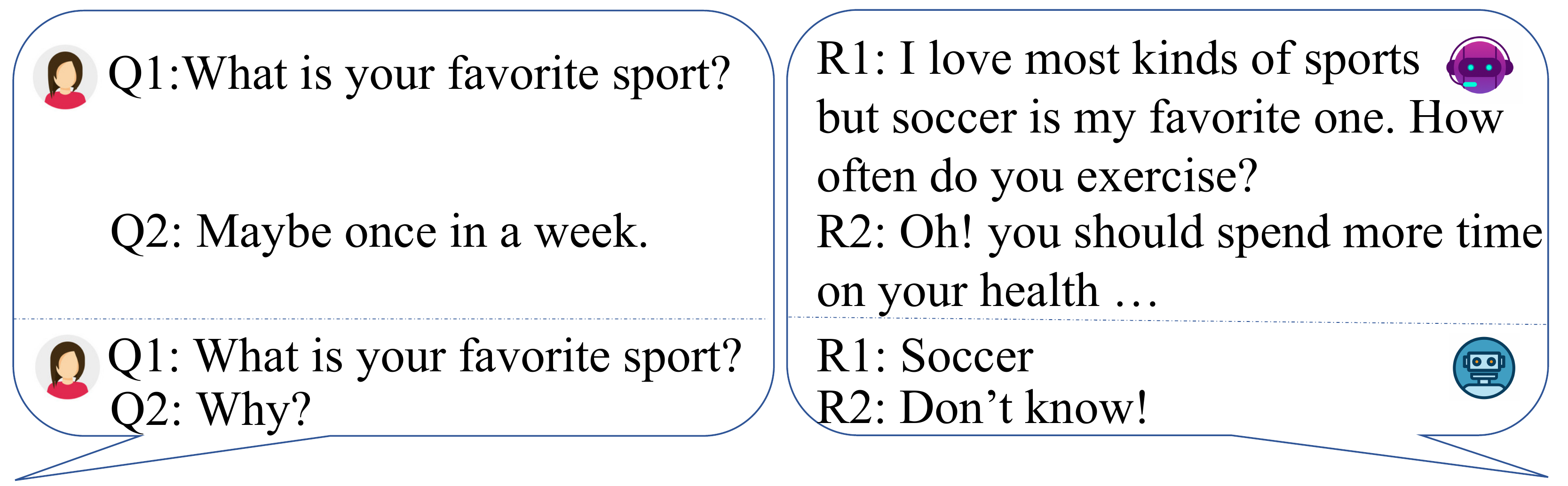}
\caption{An illustrative example of a user's conversation with two chatbots. We anticipate that the user will prefer to converse with the top chatbot because the responses are both relevant and engaging, while the bottom chatbot generates relevant but not engaging responses.}
\label{eng_exp}
\end{figure}

We argue that relevancy by itself can not capture all the characteristics of open-domain dialogue systems given their open-ended essence \cite{venkatesh2018evaluating,DBLP:journals/corr/abs-1801-03622,DBLP:conf/naacl/SeeRKW19}. For example, it is not informative to compare the output of the two dialogue systems depicted in Figure \ref{eng_exp} based on only the relevancy of generated responses, since both systems produce fairly relevant responses. Augmenting engagement scores would give a higher score to the first system, making the metric better aligned with expected user preferences.

While engagement is recognized as one of the most important metrics for open-domain dialogue evaluation~\cite{DBLP:conf/naacl/SeeRKW19,venkatesh2018evaluating,DBLP:journals/corr/abs-1801-03622}, efficient calculation of this metric poses a number of important challenges. First, existing works focus on \emph{conversation-level} engagement only, while immediate evaluation and feedback for each utterance will be more effective in providing signals to adjust the system as it proceeds in a dialogue. Second, the existing methods for evaluating engagement are mostly heuristic in nature, which can be inaccurate and usually brittle for different domains. Third, there is no systematic prior study on how well-defined and measurable {\em utterance-level} engagement is, and whether engagement measured from a single utterance can be predictive of a conversation-level engagement.   

In this paper, we propose a new proxy\footnote{The package including code, model and data can be found at  \url{https://github.com/SarikGhazarian/PredictiveEngagement}} for measuring engagement that we call {\em predictive engagement}, which, in contrast to most previous heuristics measures, operates on the utterance-level and can be learned from the data. We incorporate  predictive engagement into automatic open-domain dialogue evaluation metrics to improve the correlation with human judgements.

The contributions of the paper are four-fold:
\begin{itemize}
\item We demonstrate the feasibility of measuring {\em utterance-level} engagement by showing a high inter-annotator agreement among human annotators on rating the engagement of one query-response pair. We believe the utterance-level engagement scores can be used for real-time evaluation of dialogue systems while the conversation-level engagement can only be computed when a conversation is over. It can also be incorporated to improve the training of dialogue models.

\item We carefully study the link between utterance-level and conversation-level engagement scores and find high correlation between conversation-level engagement scores and the aggregation of individual engagement scores of a conversation's utterances. We show that assigning conversation-level engagement scores to all utterances in the same conversation is plausible, due to their high correlation. This helps us use existing resources of conversation-level engagement scores to learn utterance-level engagement scores.
\item We propose to use a transfer learning framework to leverage existing resources of conversation-level engagement annotations to build an accurate utterance-level engagement scorer for a target domain with a few additional human-annotated data.

\item Finally, we show that incorporating utterance-level predictive engagement scores into existing automatic evaluation metrics can lead to more accurate evaluation systems, which have higher correlation with human judgements. 
\end{itemize}

\section{Related Work}

The evaluation of open-domain dialogue systems is much harder than the evaluation of the task-oriented dialog systems since users do not interact with systems to achieve a specific goal. 
N-gram based evaluation metrics such as BLEU~\cite{DBLP:conf/acl/PapineniRWZ02} and ROUGE~\cite{rouge-a-package-for-automatic-evaluation-of-summaries} have poor correlation with human judgments because of the vast range of diverse valid responses in open-domain dialogue systems \cite{DBLP:conf/emnlp/LiuLSNCP16}.

Many dialogue researchers have thus resort to human evaluations to demonstrate the efficiency of their systems \cite{shang2018learning}. However, the process of gathering human judgments is neither financially nor temporally feasible, specifically for a model's hyper-parameters selection. \newcite{DBLP:conf/naacl/HashimotoZL19} brought up another shortcoming of human evaluation in assessing the response diversity and the model's generalization capability. 

\subsubsection{Learnable Evaluation Metrics}
The mentioned constraints motivate researchers to seek more accurate automatic evaluation metrics with close correlation to human judgments.
Many researchers have applied different machine learning methods such as adversarial training or classification techniques to measure the appropriateness aspect of generated responses \cite{DBLP:conf/emnlp/LiMSJRJ17,DBLP:journals/corr/KannanV17,DBLP:conf/acl/LoweNSABP17}. 

\subsubsection{Relevance Metrics}
The Referenced metric and Unreferenced metric Blended Evaluation Routine (RUBER) is an automatic evaluation metric recently proposed by \newcite{DBLP:conf/aaai/TaoMZY18} that combines relevancy score of a response to a given query with its similarity to the ground-truth response. Their proposed neural-based model trains the relevancy score of each utterance based on negative sampling, while the referenced metric measures cosine similarity of ground-truth and generated response vectors. \newcite{ghazarian2019better} improved RUBER by incorporating contextualized BERT embeddings \cite{DBLP:journals/corr/abs-1810-04805} into both referenced and unreferenced metrics. Throughout this paper, we will call this model contextualized RUBER. Although, RUBER and its improved version have high correlation with human judgements, they both consider only the relevancy metric, which is not adequate for fair evaluation of open-domain dialogue systems as demonstrated in Figure \ref{eng_exp}.

\subsubsection{Engagement Metrics}

Engagement is a substantial metric that shows user willingness to continue conversing with the system \cite{DBLP:conf/interspeech/YuAW04,ma2018towards,inoue2018engagement} and has been studied in the context of dialogue systems \cite{yu2016wizard,DBLP:conf/acl/KielaWZDUS18,DBLP:conf/naacl/SeeRKW19}. Many researchers have considered engagement as a useful metric toward achieving better dialogue systems \cite{yu2016wizard,DBLP:conf/acl/KielaWZDUS18}. PERSONACHAT dataset, which includes persona information, has been prepared by \newcite{DBLP:conf/acl/KielaWZDUS18} with the focus on having more engaging chatbots. \newcite{yu2016wizard} argued that optimizing open-domain dialogue systems only on relevancy is not enough and engagement can improve the quality of these systems. 
In these efforts, users and experts have been asked to annotate the engagement score of utterances. \newcite{DBLP:conf/naacl/SeeRKW19} have framed human opinion about overall quality of dialogue systems with two main metrics; humanness and engagingness. They have studied how controlling various attributes such as repetition, specificity and question-asking leads to higher engaging responses. 

Engagement estimation has been addressed in many spoken dialogue systems based on a listener's multimodal behavior or acoustic features of conversations \cite{DBLP:conf/interspeech/YuAW04,inoue2018engagement}.
Heuristic measurements of engagement scores have been proposed by many researchers, but have their own shortcomings \cite{venkatesh2018evaluating,khatri2018alexa,DBLP:journals/corr/abs-1906-09308}. In the Alexa prize competition, the engagement score of dialogue systems is calculated based on the number of turns and the total duration of conversation \cite{venkatesh2018evaluating,khatri2018alexa}. This approach suffers from the weakness that it may classify a long conversation as engaging whereas two interlocutors were simply having difficulty understanding each other. In addition, this evaluation has to wait until the end of the conversation to estimate engagement.  

\newcite{DBLP:journals/corr/abs-1906-09308} considered a dialogue system engaging when it has the ability to ask questions during a conversation and generate longer responses. There are many counter examples for these metrics such as long responses that do not make sense or dialogue systems that do not ask questions but are still capable of generating interesting responses. As a result, they failed to show these metrics have high correlations with human judgements.

\newcite{DBLP:journals/corr/abs-1904-13015} applied automatic evaluation metrics to enhance the quality of responses generated by dialogue systems. They did not directly train a model to predict the engagement score, rather they asked annotators about interestingness and willingness to continue the conversation. They used the answers to these two questions as a proxy for engagement, which required additional human annotations. 

\section{Analysis of Engagement Scores}

We propose to build a learnable model for utterance-level engagement scores to evaluate open-domain dialogs. In this section, we discuss utterance-level and conversation-level engagement scores and investigate their connections.
\subsection{Conversation-level Engagement Scores}
Engagement is defined as a user's inclination to continue interacting with a dialogue system \cite{inoue2018engagement,ma2018towards}. 
In many existing chatbot competitions like NeurIPS ConvAI \footnote{\url{http://convai.io/2017/data/}} and
Amazon Alexa prize \footnote{\url{https://developer.amazon.com/alexaprize}},
users are asked to evaluate whole conversations based on how engaging and attractive they are in maintaining interaction. We define this as conversation-level engagement scores. In this work, we utilize the ConvAI dataset since it is publicly accessible. 

In human evaluation rounds of ConvAI competition, participants and volunteers conversed with a human or a chatbot via Telegram and Facebook messaging services, where their peers had been randomly assigned to them \cite{logacheva2018dataset}. From overall 4750 dialogues, the majority of conversations were between a human and a bot and 526 were human-to-human conversations. The interlocutors, participants and chatbots, rated utterances as well as conversations on different conversational aspects, where engagement was collected at the conversation-level in the range of 0 to 5 (0 as not engaging at all and 5 as extremely engaging). Engagement scores for human-to-human conversations were calculated by averaging user ratings, while for human-to-bot conversations, only the human's opinion was used as a dialogue's engagement score.
The first row in Table \ref{convai_stats} demonstrates the distribution of the conversations with different engagement scores.  

\begin{table}[t]
\small
\begin{center}
\begin{tabular}{lcccccc}
\toprule
  &\multicolumn{6}{c}{\textbf{Engagement Scores}}  \\
  \cmidrule{2-7}
  & \textbf{0} & \textbf{1} &\textbf{2} & \textbf{3}  & \textbf{4}  & \textbf{5}\\\midrule
  \textbf{Conversations} &  1690 & 21 & 81 & 47 & 147 & 63 \\
  \hline
  \textbf{Utterances} & 10122 & 45 & 238 & 444 & 1492 & 783 \\
  \bottomrule 
\end{tabular}
\end{center}
\caption{Data statistics of the ConvAI evaluation dataset. The first row shows conversations with their corresponding engagement scores extracted from the original ConvAI dataset; the second row contains the number of utterances and their engagement scores automatically assigned by our heuristics.}
\label{convai_stats}
\end{table}

\subsection{Utterance-level Engagement Scores}
To explore the efficiency of incorporating engagement into existing successful automatic evaluation metrics measuring relevancy at the utterance level \cite{DBLP:conf/aaai/TaoMZY18,ghazarian2019better}, we need to investigate whether or not an engagement score can be measured at the utterance level. 
We propose to first study whether humans are capable of scoring engagement of a response for a given query without knowing any context or previous utterances. To achieve this, we executed experiments to check users' agreement level about engagement of each utterance. We conducted Amazon Mechanical Turk (AMT) experiments on randomly selected 50 conversations from ConvAI, 25 human-to-human and 25 human-to-bot dialogues.
Overall 297 utterance pairs have been extracted and rated by annotators in the same range (1-5) of engagement score in ConvAI. 49 workers participated in about 215 surveys, where each utterance pair has been annotated by 5 individual workers. We rejected users that did not pass attention-check tests in the surveys and reassigned their pairs to other workers. 
Eventually, as Table \ref{uttagg_user_anno} demonstrates, the mean $\kappa$ agreement and mean Pearson correlation between evaluators participating in our experiments were 0.52 and 0.93. In the context of dialogue system evaluation where agreement is usually quite low \cite{venkatesh2018evaluating,DBLP:journals/corr/abs-1906-09308,DBLP:journals/corr/abs-1904-13015}, these numbers show relatively high agreement between annotators. This provides evidence that engagement can be measured not only at the conversation level but also at the utterance level.

\begin{table}[t]
\small
\begin{center}
\begin{tabular}{cccc}
 \toprule
  \textbf{Utterances} & \textbf{Annotators} & \textbf{Kappa Agreement} & \textbf{Pearson}\\\midrule 
  297 & 49 & 0.52 & 0.93 \\
  \bottomrule 
\end{tabular}
\end{center}
\caption{The results for the Amazon Mechanical Turk (AMT) experiments on utterance-level engagement. 49 annotators annotated 297 utterances and demonstrated quite high inter-annotator Kappa agreement and Pearson correlation between annotations.}
\label{uttagg_user_anno}
\end{table}

\subsection{Utterance-level and Conversation-level Engagement Scores}
\begin{figure}[t]
    \centering
    \includegraphics[width=\linewidth]{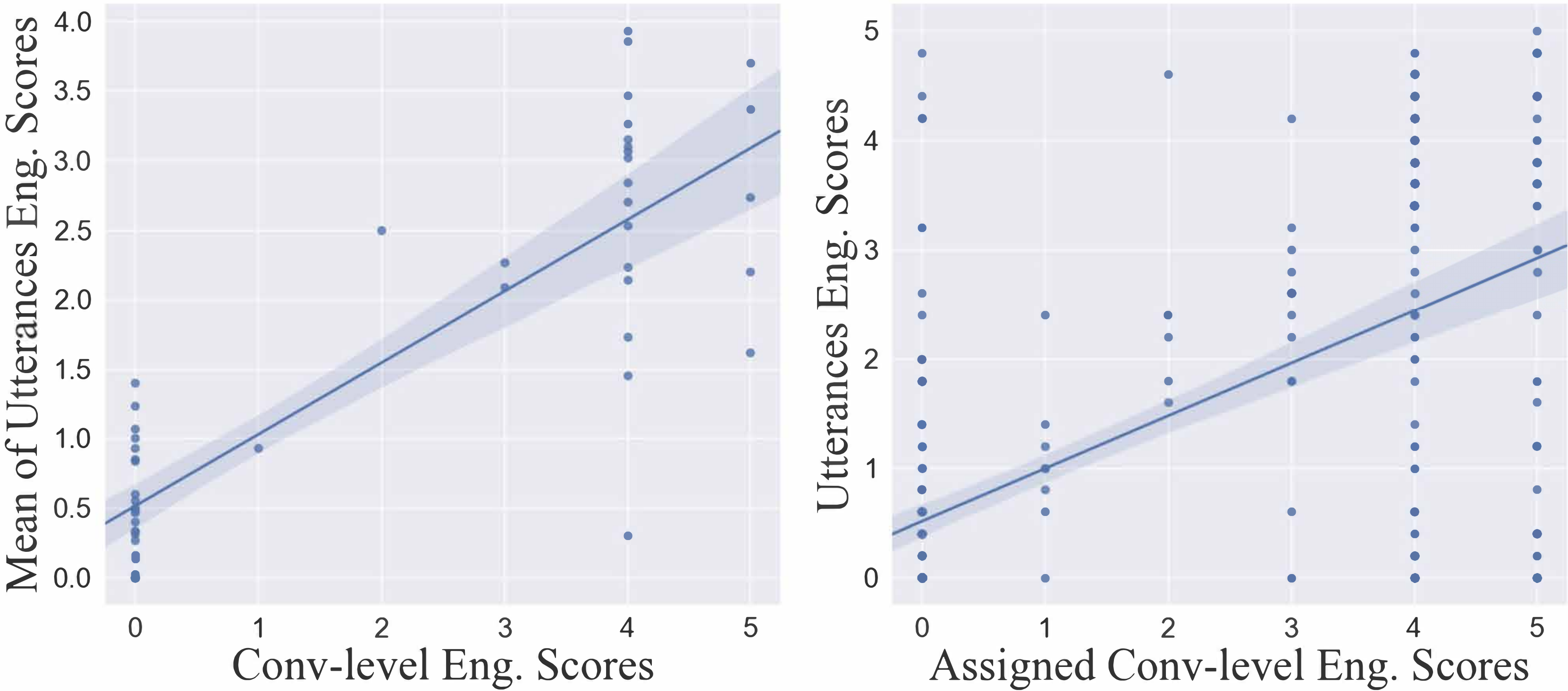}
    \caption{The left scatterplot depicts the correlation between the ground-truth conversation-level engagement scores and the mean aggregation of engagement scores of utterances for 50 conversations conducted in AMT experiment. The Pearson correlation value is 0.85. The right scatterplot depicts the correlation between the engagement scores of 297 utterances annotated by human in AMT experiment and heuristically assigned conversation-level engagement score to all utterances in the conversation. The Pearson correlation value is 0.60.}
    \label{uttconv_con}
\end{figure}

The high inter-annotator agreement on utterance-level engagement scores motivated us to study if conversation-level engagement scores can be converted into utterance-level ones. This can be beneficial as we can leverage them to train utterance-level engagement scorer since there are no prior datasets for utterance-level engagement scores. The scorer can later be incorporated into existing automatic evaluation metrics.
Hence, we ask the following research questions:
\begin{itemize}

\item{} \textbf{Is there a high correlation between the aggregated utterance-level engagement scores and conversation-level engagement score?} For this purpose, we used the engagement scores annotated by AMT workers for 297 utterances of ConvAI dataset, where each utterance's engagement score was the average of five individual annotators ratings. In order to calculate the intended correlation, we considered the engagement score of each conversation as the ground-truth and aggregated its utterances' engagement scores annotated by AMT workers to get the predicted conversation engagement score. Table \ref{uttagg_conv_corr} shows the computed correlations using different aggregation methods. The highest correlation is based on mean aggregation of utterance-level engagement scores, which is presented in the left scatterplot of Figure \ref{uttconv_con}.
Considering minimum or maximum aggregation of engagement scores for utterances as the conversation's overall score leads to lower correlation since not only all utterances of a good conversation are not engaging but also all utterances of a bad conversation are not boring. 

\begin{table}[t]
\begin{center}
\begin{tabular}{cc}
 \toprule
  \textbf{Aggregation Method} &\textbf{Pearson Correlation}\tiny(p-value)\\\midrule 
  Min & 0.49 \tiny(\textless3e-4) \\
  Max & 0.72 \tiny(\textless4e-9)\\
  Mean & \bf 0.85 \tiny(\textless9e-15) \\
  \bottomrule 
\end{tabular}
\end{center}
\caption{The Pearson correlation between engagement scores of 50 randomly selected conversations from ConvAI and the aggregated engagement scores of their utterances annotated by AMT workers with different aggregation methods.}
\label{uttagg_conv_corr}
\end{table}

\item{} \textbf{Is there a high correlation between utterance-level engagement scores and conversation-level engagement scores assigned to all utterances in the conversation?} In this part, we assigned the ConvAI conversation-level engagement scores to each of its utterances and then computed the Pearson correlation between these assigned scores and the scores from AMT workers. The computed Pearson correlation was 0.60, a relatively high correlation that has been depicted in right scatterplot of Figure \ref{uttconv_con}. 
There are cases where the difference between human ratings and assigned scores is clearly visible. 
Even though there are these mismatches, there is no  publicly available dataset containing utterance-level engagement scores. The relatively high correlation between these scores enabled us to assign conversation-level scores to all utterances in the ConvAI dataset and used it for further experiments. The second row in Table \ref{convai_stats} shows these utterances with their assigned engagement scores.
\end{itemize}

As is shown in Table \ref{convai_stats}, the majority of utterances have zero engagement scores and the remaining are accumulated near labels 4 and 5. Therefore, we split the range of engagement scores from 1 to 5 into a binary range (considering all scores less than or equal to 2 as not engaging and greater than 2 as engaging); around 80 percent of the utterances are labeled as not engaging, and the remaining as engaging.

\section{Engagement Classifier}

As there is an absence of baseline models for automatically measuring utterance-level engagement scores, we consider one feature-based model and one neural-based model as baselines.

\begin{figure}[t]
\centering
\includegraphics[width=\linewidth]{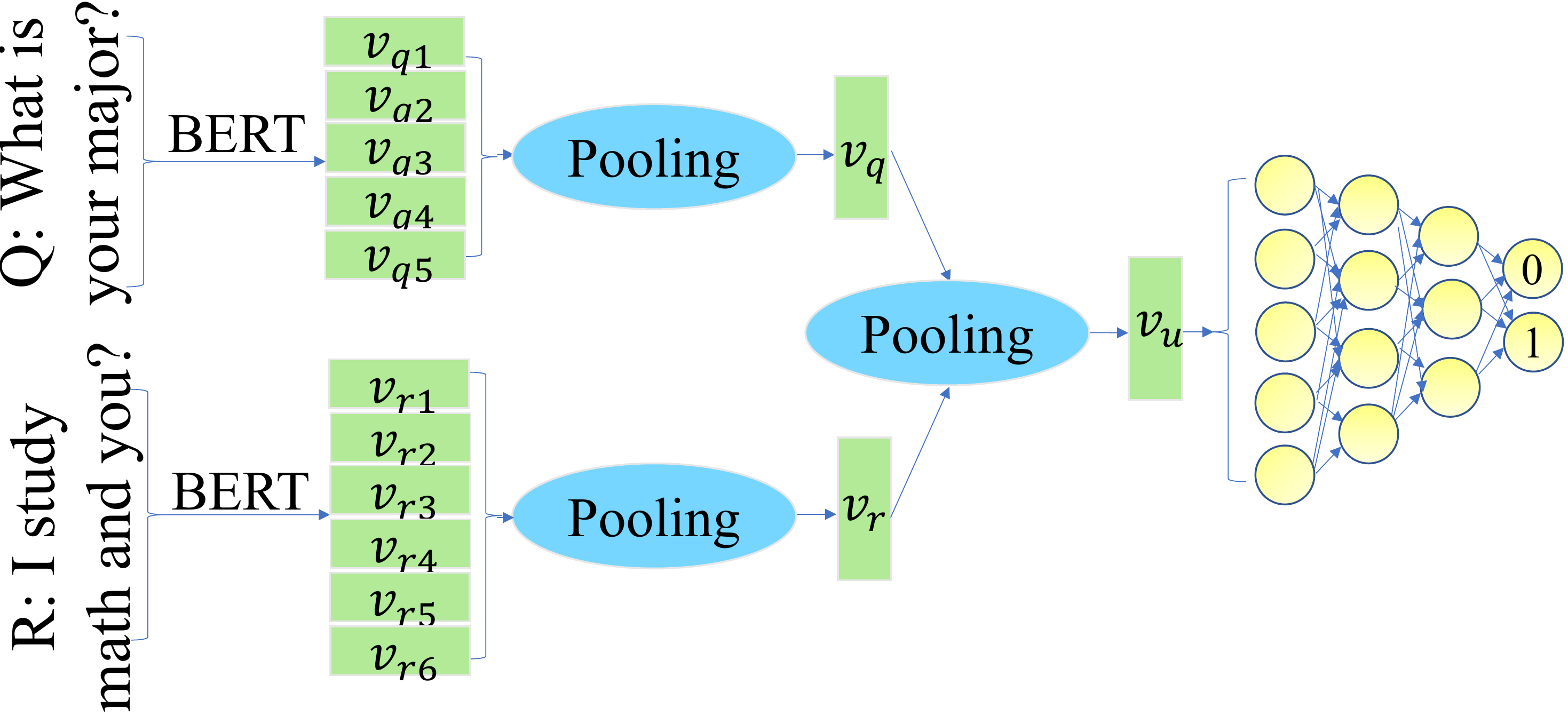}
\caption{An illustration of the proposed utterance-level engagement classifier.}
\label{fig:eng_exp}
\end{figure}

\begin{itemize}
\item The feature-based model is an SVM classifier with a predefined set of features including n-grams, length of each response and number of distinct words in each response.
\item The neural-based model is a classifier with static word2vec embeddings as input and two Bidirectional Recurrent Neural Networks (Bi-RNNs) to map words embeddings into vector representations for both query and response, with a Multilayer Perceptron (MLP) classifier on top of the concatenated vector of each utterance pair.
\item Our proposed engagement classifier is shown in Figure \ref{fig:eng_exp}. It takes a pair of query and response as input and classifies it as engaging or not engaging. 
We choose to use BERT~\cite{DBLP:journals/corr/abs-1810-04805} embeddings as input to our model since~\newcite{ghazarian2019better} showed superior results using BERT to evaluate the relevance of a response. 
The utterance vectors are computed by simply taking the max or mean pooling of their contextualized word embeddings. This works because these embeddings are computed by pretrained deep bidirectional transformers that already have information of the context. \newcite{ghazarian2019better} showed simple pooling strategy worked better than adding an additional BiLSTM layer when computing relevance. 
The utterance vectors of query and response pairs are then passed through an MLP classifier with cross entropy loss to classify the utterance as 0 (not engaging) or 1 (engaging).
\end{itemize}

\section{Experimental Setup and Results} 
In the experiments, we explored the efficacy of augmenting engagement scores inferred by our proposed model on open-domain dialog evaluation. We trained our proposed model on the ConvAI dataset and then fine-tuned it on the Daily Dialogue dataset that we used to evaluate the performance of automatic evaluation metrics. We augmented predicted engagement scores with relevancy scores from baseline models and examine their correlation with human judgements.  
\subsection{Baseline Models}
In order to study the efficacy of combining engagement scores with existing evaluation metrics for dialogue systems evaluation, we used the unreferenced scores in RUBER~\cite{DBLP:conf/aaai/TaoMZY18} and Contextualized RUBER~\cite{ghazarian2019better} as the baseline metrics. In our experiments, we did not consider the referenced metric that measures the similarity between generated responses with references since~\newcite{ghazarian2019better} showed that considering only the unreferenced scores yielded higher correlation with human judgements. 
The unreferenced score proposed by \newcite{DBLP:conf/aaai/TaoMZY18} is computed by an MLP neural model which is trained with a ranking loss. This loss function maximizes the inferred score between positive and negative samples which are obtained from dataset and randomly matched query and response pairs, respectively. 
For the Contextualized RUBER baseline, we considered the best model proposed by \newcite{ghazarian2019better} which is an MLP classifier that takes contextualized embeddings as richer representation of words and uses cross-entropy loss function.

\subsection{Datasets}
In order to explore the effect of engagement score on existing automatic evaluation metrics including RUBER and contextualized RUBER, we needed a dataset to train the proposed engagement classifier and a dataset to train an automatic dialogue evaluation metric to compare with the baselines. We used the ConvAI dataset for the first purpose since it has annotations for conversation-level engagement scores. We used the Daily Dialogue dataset to evaluate the efficiency of the utterance-level engagement scores for open-domain dialogue evaluation.

\subsubsection{ConvAI }
To train the utterance-level engagement model, we used the engagement scores of conversations in ConvAI assigned to 13,124 utterance pairs as input data shown in the second row of Table \ref{convai_stats}. We split this dataset into 60/20/20 parts as train/validation/test sets. Table \ref{convai_eng_stats} shows these sets with the number of utterances labeled as 0 or 1.

\subsubsection{Daily Dialogue Dataset}
The Daily Dialog dataset \footnote{\url{http://yanran.li/dailydialog}} is an open-source multi-turn open-domain dialogue dataset that includes daily conversations between humans on different topics. We used a part of this dataset including 22,000/1,800/2,100 pairs of train/test/validation sets for training the relevancy score of RUBER and contextualized RUBER as baselines models. In order to explore the effects of engagement scores on automatic evaluation metrics, we used the following datasets. In subsequent sections, we refer to each dataset based on specified names.
\begin{itemize}
\item \textbf{300 utterances with generated replies}: this is a human annotated dataset\footnote{\url{http://vnpeng.net/data/DailyDialog_annotated.zip}} about the quality of 300 utterance pairs randomly selected from the test set of the Daily Dialogue dataset released by \newcite{ghazarian2019better}, where replies are generated based on an attention-based sequence-to-sequence dialogue model. 
\item \textbf{300 utterances with human-written replies}: Most replies in the above mentioned dataset are completely off-topic and do not make sense; therefore the engagement score will not add extra information about them. In order to have a fair assessment of successful dialogue systems that mainly include relevant responses, we repeated the experiments done by \newcite{ghazarian2019better} on the same 300 queries but with their ground-truth responses that mostly are relevant but not always engaging. We asked evaluators to judge each response's overall quality in the range of 1 to 5 (low quality to high quality). Each pair is annotated by 3 individual workers; overall 24 annotators contributed in this experiment.
\end{itemize}

\begin{table}[t]
\begin{center}
\begin{tabular}{ccc}
 \toprule
  & \textbf{Engagement = 0} & \textbf{Engagement = 1} \\\hline
  \textbf{Train} &  6222 & 1562 \\
  \hline
  \textbf{Validation} & 2121 & 575 \\
  \hline
  \textbf{Test} &2062 & 582 \\
  \bottomrule 
\end{tabular}
\end{center}
\caption{ConvAI train/valid/test sets of utterances with their engagement score labels}
\label{convai_eng_stats}
\end{table}

\begin{figure}[t]
\centering
\includegraphics[width=.7\linewidth]{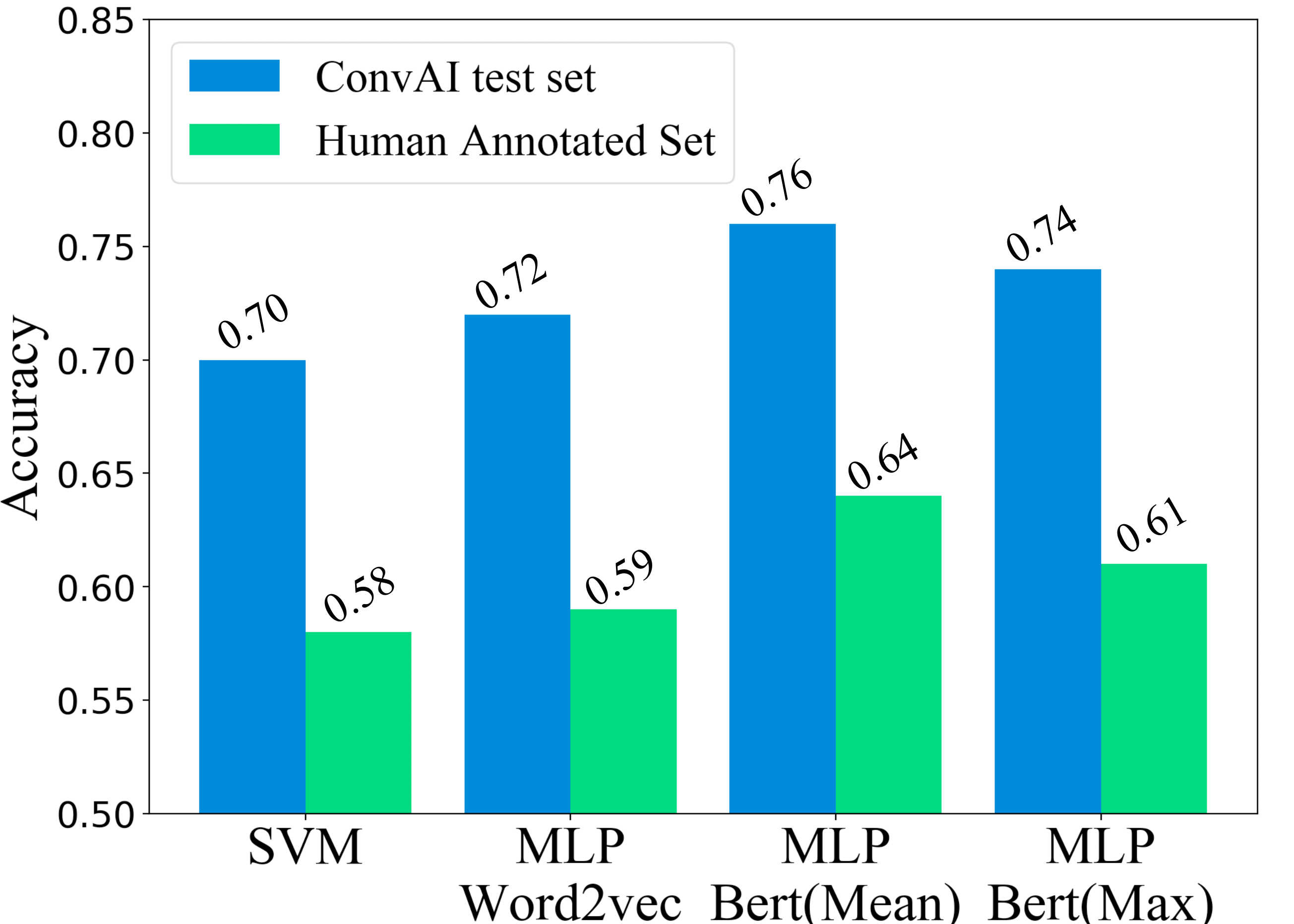}
\caption{Balanced accuracy (ROC AUC) of different utterance-level engagement classifiers on the ConvAI test set (the last row in Table \ref{convai_eng_stats}) and human annotated test set (Table \ref{uttagg_user_anno}). The first two groups of bars show SVM and MLP classifier performance based on word2vec embeddings; the remaining bars are our proposed classifiers based on BERT embeddings with mean and max pooling strategies.}
\label{eng_acc}
\end{figure}

\begin{figure}[t]
\centering
\includegraphics[width=\linewidth]{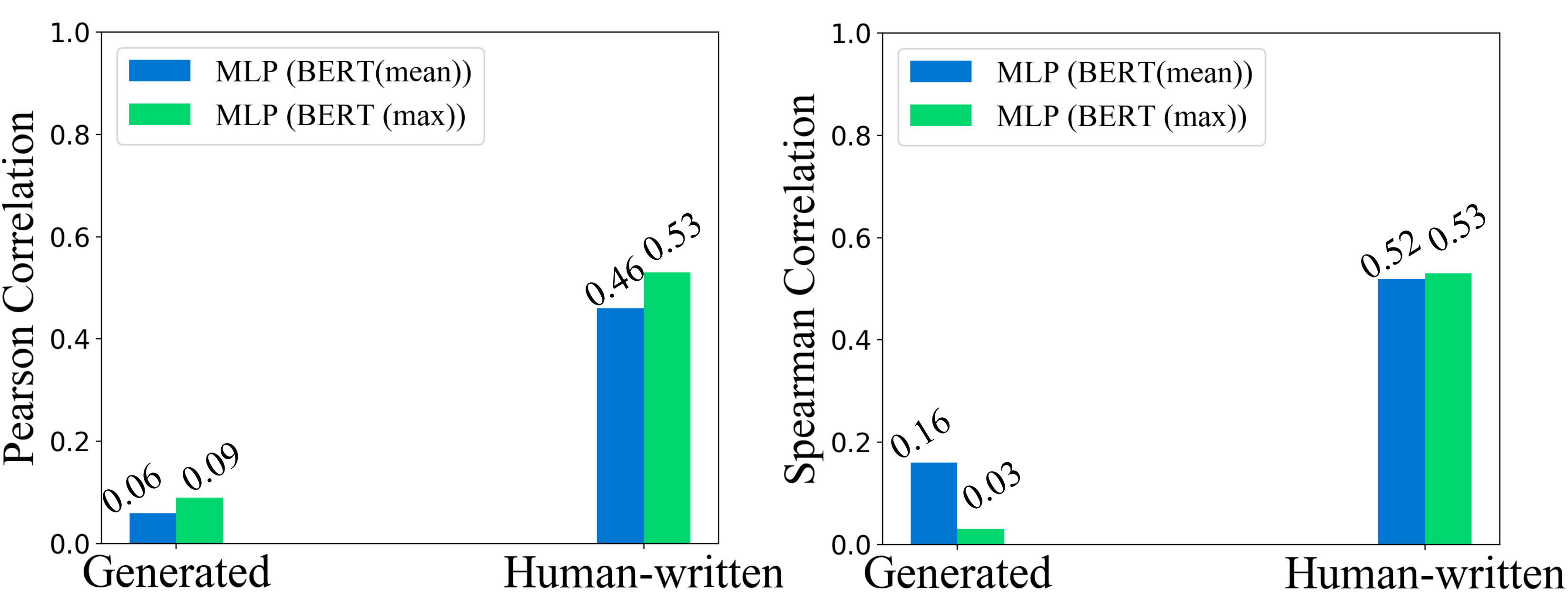}
\caption{Pearson and Spearman correlations between predictive engagement scores and human judgements for generated and human-written responses.}
\label{rel_nrel_corr}
\end{figure}

\begin{table*}[t]
\begin{center}
\begin{tabular}{cccc}
 \toprule
 \textbf{Dataset} &\textbf{Metric} & \bf Pearson & \bf Spearman  \\\midrule
 \multirow{6}{*}{300 Generated Responses} & RUBER\_relavance & 0.28 & 0.30\\
  & Ctx\_RUBER\_relevance & 0.55 & 0.45\\
  & MLP BERT(mean) & 0.06 & 0.16\\
  & MLP BERT(max) & 0.09 & 0.03\\
  & MLP BERT(mean) + Ctx\_RUBER\_relevance & 0.48 & 0.48\\
  & MLP BERT(max) + Ctx\_RUBER\_relevance & 0.52 & 0.44\\
  \midrule 
  \multirow{5}{*}{300 Human-written Responses} & RUBER\_relavance & 0.04 & 0.02\\
  & Ctx\_RUBER\_relevance & 0.14 & 0.12\\
  & MLP BERT(mean) & \bf 0.46 & \bf 0.52\\
  & MLP BERT(max) & \bf 0.53 & \bf 0.53\\
  & MLP BERT(mean) + Ctx\_RUBER\_relevance & \bf 0.36 & \bf 0.39\\
  & MLP BERT(max) + Ctx\_RUBER\_relevance & \bf 0.32 &  \bf 0.39\\
  \midrule 
  \multirow{5}{*}{\begin{minipage}{0.5\columnwidth} 600 Generated and Human-written Responses \end{minipage}} & RUBER\_relavance & 0.24 & 0.30\\
  & Ctx\_RUBER\_relevance & 0.54 & 0.55\\
  & MLP BERT(mean) & 0.39 & 0.45\\
  & MLP BERT(max) & 0.35 & 0.36\\
  & MLP BERT(mean) + Ctx\_RUBER\_relevance & \bf 0.61 & \bf 0.64\\
  & MLP BERT(max) + Ctx\_RUBER\_relevance & \bf 0.60 & \bf 0.62\\
  \midrule 
\end{tabular}
\end{center}
\caption{Pearson and Spearman correlations between human judgements and several automatic dialogue evaluation metrics on generated responses, human-written responses, and their mixture. We adopt the mean aggregation of the relevance score of contextualized RUBER mentioned as Ctx\_RUBER and the predictive utterance-level engagement scores. The first two rows in each group show correlations between human judgements and baseline models with only relevance scores, the middle two rows are for only engagement scores and the last two rows add engagement scores into relevance scores. Boldface indicates the improvements are significant compared to the baseline in the corresponding group (p $< .05$).}
\label{engrel_results}
\end{table*}

\begin{table*}[t]
\begin{center}
\begin{tabular}{llcccc}
 \toprule
  \textbf{Query} &\textbf{Response}  & \bf \begin{minipage}{0.1\columnwidth}\centering Human \end{minipage}& \bf \begin{minipage}{0.1\columnwidth}\centering Rel. \end{minipage} & \bf \begin{minipage}{0.1\columnwidth}\centering Eng.\end{minipage} & 
  \bf \begin{minipage} {0.1\columnwidth}\centering Rel. \\\centering + \\ \centering Eng.\end{minipage}\\
  \midrule 

  \begin{minipage}{0.6\columnwidth}OK. What's the reason you are sending her flowers? \end{minipage}&\begin{minipage}{0.6\columnwidth} Today's her birthday and she told me she wants me to buy her flowers.  \end{minipage}& 0.92 & 0.99 & 0.88 & 0.94 \\
    \midrule 
  \begin{minipage}{0.6\columnwidth}The kitchen may be large, but it doesn't have any storage space. \end{minipage}&\begin{minipage}{0.6\columnwidth} The master suite is supposed to be quite elegant. Maybe it will be a little better. \end{minipage}& 0.75 & 0.65 & 0.94 & 0.80 \\
   \midrule 
  \begin{minipage}{0.6\columnwidth}Not long, because people rush for lunch. \end{minipage}&\begin{minipage}{0.6\columnwidth}The line sure does move fast.\end{minipage}& 0.33 & 0.82 & 0.11 & 0.47 \\
  \midrule 
  \begin{minipage}{0.6\columnwidth}That's a good idea. And remind them to be slow at the beginning, not to run into the railings. \end{minipage}&\begin{minipage}{0.6\columnwidth}OK. Anything else?\end{minipage}& 0.42 & 0.84 & 0.14 & 0.49 \\
  \bottomrule 

\end{tabular}
\end{center}
\caption{Examples of evaluation scores for utterances from the Daily Dialogue Dataset. We used unreferenced score of Contextualized RUBER as relevance score and MLP BERT(mean) as engagement score, where the numbers are rounded into 2 digits. The incorporation of engagement scores into relevance scores yields scores closer to human judgements -- the main goal of automatic evaluation metrics.}
\label{examples_utt_eval}
\end{table*}
\subsection{Implementation Details} We trained our proposed model for utterance-level engagement score along with two baseline models on ConvAI dataset. Due to the imbalanced nature of this dataset, we used a weighted loss function for training purposes and balanced accuracy scores for evaluation.
We trained the SVM classifier with a linear kernel function and 0.1 C parameter. Word2vec embeddings used in the neural baseline classifier are 300 dimensional embeddings trained on about 100 billion words of the Google News Corpus \cite{DBLP:journals/corr/abs-1301-3781}. The baseline neural model is a one layer MLP classifier with tanh as the activation function, a learning rate of $10^{-5}$ and 0.8 dropout rate. Our proposed model uses BERT 768 dimensional vectors pre-trained on the Books Corpus and English Wikipedia as words embeddings \cite{DBLP:journals/corr/abs-1810-04805}. The model is trained with a weighted cross entropy loss function. The MLP classifiers are 3-layer networks with 64, 32 and 8 hidden units. Learning rate in the MLP classifier based on mean pooling of word embeddings is $10^{-3}$, while with max pooling  it is $10^{-2}$.
The performance of all trained models has been demonstrated in Figure \ref{eng_acc}. The blue bars show the balanced accuracy of models on the ConvAI test set (Table \ref{convai_eng_stats}), while the green bars show the balanced accuracy on utterance pairs of 50 conversations annotated by AMT workers as another test benchmark. According to results from Figure \ref{eng_acc}, our proposed models based on BERT embeddings perform better in terms of accuracy, which will be used for inferring engagement scores of utterances in the Daily Dialogue dataset.

\subsubsection{Transfer Learning} After training utterance-level engagement classifiers, we fine-tuned them on a small set of utterance pairs randomly selected from the Daily Dialog dataset excluding the pairs in 300 utterances for assessing automatic evaluation metrics. Indeed, the ConvAI dataset that the engagement models are trained on  is the source domain, and the selected dataset for fine tuning is the target domain. We recruited about 45 participants from AMT to annotate 300 pairs from Daily Dialog dataset as engaging or not engaging. Around half of the selected pairs were from Daily Dialogue queries and their ground truth responses that mostly are part of engaging conversations between two humans. The other half were queries and responses generated by attention-based sequence-to-sequence dialogue system that mostly were not engaging. We attained a mean $\kappa$ agreement of 0.51 between users that passed the attention-check tests attached to AMT surveys.

\subsection{Experimental Results}
Performance of automatic evaluation metrics for open-domain dialogue systems are measured based on their correlation with human judgements. Higher correlations indicate these metrics can be a great substitution for human evaluations.

\subsubsection{Quantitative Results} We inferred the engagement scores from fine-tuned utterance-level engagement models for the 300 utterances with generated replies and aggregated them with the relevance scores obtained from the Contextualized RUBER model. We only included the mean aggregation of relevance and engagement metrics that resulted in the highest correlation in comparison with the other two aggregations (minimum and maximum) that we tried. 
Each part of the table \ref{engrel_results} illustrates the correlations between human judgements with relevance, engagement and the combination of these two metrics respectively.
As is shown in the first part of Table \ref{engrel_results}, the correlations between human judgements and the two evaluation metrics are very close to the baseline metrics that only compute relevance scores. Many off-topic replies generated by the attention-based sequence-to-sequence dialogue system could be the reason for this observation. 
According to the second part of Table \ref{engrel_results}, the Pearson and Spearman correlations between human judgments and the relevance scores for the 300 utterances with human-written replies is low. Incorporating engagement scores leads to higher correlations with human judgements. Indeed, the baseline models score the majority of human-written responses very high, while users consider other aspects such as engagement for giving the utterance an overall quality score. 

Figure~\ref{rel_nrel_corr} more clearly depicts that the correlation between human judgements and engagement-only scores on the 300 utterances with generated replies is low. This is probably because the annotators do not pay attention to other aspects like engagement for evaluating a response that is not relevant to a given query.
Figure \ref{rel_nrel_corr} also illustrates the positive effect of considering engagement scores in evaluating the human-written responses.

We combined the two sets from the Daily Dialogue dataset, and the last part in Table \ref{engrel_results} shows the correlations on the combined 600 query-reply pairs. The higher correlations between human annotations with relevance and engagement scores illustrate the success of applying engagement as an extra score to baseline metrics in order to have a better automatic evaluation system. 

\subsubsection{Significance Test} To see whether the improvements of the correlation is statistically significant, we applied hypothesis testing to compare the dependant correlations with overlapping variables; in our case the human judgements \cite{diedenhofen2015cocor}.
According to hypothesis testing, the probability of the null hypothesis, which states that two correlations are equal or not that much different is $\leq 0.05$; thus, the improvement is significant.
 
\subsubsection{Case Study} Some real examples from the Daily Dialogue dataset are shown in Table \ref{examples_utt_eval}, which demonstrates the positive influence of aggregating engagement score with relevance score in order to have much closer evaluations to human judgements.

\section{Conclusion and Future work}

In this paper, we hypothesized that it is not adequate to compare open-domain dialogue systems solely based on the relevance of the responses. An utterance-level engagement measurement can be built to improve the automatic open-domain dialog evaluation metrics. To this end, we verified the feasibility of measuring utterance-level engagement and showed a high correlation between the utterance-level and conversation-level engagement scores.
We incorporated the utterance-level engagement scores inferred by our proposed model into other relevance-based evaluation metrics and showed an improved correlation to human judgements. 
We plan to apply our proposed automated engagement metric to guide the training of a dialogue system to encourage the generation of more interesting and engaging responses.

\section{Acknowledgments}
This work is funded by the Contract W911NF-15-1-0543 with the US Defense Advanced Research Projects Agency (DARPA). We thank the anonymous reviewers for their useful comments and the members of the USC/ISI PLUS lab for their constructive feedback.

\bibliographystyle{aaai}
\bibliography{references}
\end{document}